\documentclass[lettersize,journal]{IEEEtran}
\usepackage{amsmath,amsfonts}
\usepackage{array}
\usepackage[caption=false,font=normalsize,labelfont=sf,textfont=sf]{subfig}
\usepackage{textcomp}
\usepackage{stfloats}
\usepackage{url}
\usepackage{verbatim}
\usepackage{graphicx}
\usepackage{cite}
\usepackage[ruled,linesnumbered]{algorithm2e}
\usepackage{siunitx}
\SetKwInput{KwInit}{Initialize}
\SetKwInput{KwInst}{Instantiate}
\usepackage[hidelinks]{hyperref}

\begin{document}

\title{EnsembleFollower: A Hybrid Car-Following Framework Based On Reinforcement Learning and Hierarchical Planning}

\author{Xu Han, Xianda Chen,~\IEEEmembership{Student Member,~IEEE}, Meixin Zhu,~\IEEEmembership{Member,~IEEE}, Pinlong Cai, Jianshan Zhou, Xiaowen Chu,~\IEEEmembership{Senior Member,~IEEE}
\thanks{This study is partly supported by Guangzhou Basic and Applied Basic Research Project (2023A03J0106) and Guangzhou Municipal Science and Technology Project (2023A03J0011).}
\thanks{Xu Han and Xiaowen Chu are with the Data Science and Analytics Thrust, Information Hub, The Hong Kong University of Science and Technology (Guangzhou), Guangzhou, China (email: xhanab@connect.ust.hk, xwchu@ust.hk).}%
\thanks{Xianda Chen and Meixin Zhu are with the Intelligent Transportation Thrust, Systems Hub, The Hong Kong University of Science and Technology (Guangzhou), Guangzhou, China; also with Guangdong Provincial Key Lab of Integrated Communication, Sensing and Computation for Ubiquitous Internet of Things (email: xchen595@connect.hkust-gz.edu.cn, meixin@ust.hk).}%
\thanks{Pinlong Cai is with the Shanghai Artificial Intelligence Laboratory, Shanghai, China (email: caipinlong@pjlab.org.cn).}%
\thanks{Jianshan Zhou is with State Key Laboratory of Intelligent Transportation Systems, Beijing Key Laboratory for Cooperative Vehicle Infrastructure Systems \& Safety Control, and School of Transportation Science and Engineering, Beihang University, Beijing 100191, China (jianshanzhou@foxmail.com).}%
\thanks{\textit{(Corresponding authors: Xiaowen Chu, Meixin Zhu).}}%
}



\maketitle

\begin{abstract}
Car-following models have made significant contributions to our understanding of longitudinal driving behavior. However, they often exhibit limited accuracy and flexibility, as they cannot fully capture the complexity inherent in car-following processes, or may falter in unseen scenarios due to their reliance on confined driving skills present in training data. It is worth noting that each car-following model possesses its own strengths and weaknesses depending on specific driving scenarios. Therefore, we propose EnsembleFollower, a hierarchical planning framework for achieving advanced human-like car-following. The EnsembleFollower framework involves a high-level Reinforcement Learning-based agent responsible for judiciously managing multiple low-level car-following models according to the current state, either by selecting an appropriate low-level model to perform an action or by allocating different weights across all low-level components. Moreover, we propose a jerk-constrained kinematic model for more convincing car-following simulations. We evaluate the proposed method based on real-world driving data from the HighD dataset. The experimental results illustrate that EnsembleFollower yields improved accuracy of human-like behavior and achieves effectiveness in combining hybrid models, demonstrating that our proposed framework can handle diverse car-following conditions by leveraging the strengths of various low-level models.
\end{abstract}


\section{Introduction}
\IEEEPARstart{A}{s} a predominant driving scenario, car following plays a crucial role in the overall performance and safety of autonomous driving systems. The primary objective of car-following models is to effectively manage vehicle speed in order to maintain secure and comfortable following distances. By advancing research in autonomous car-following control algorithms, we aim to alleviate human drivers' cognitive burden, enhance road safety, and optimize traffic flow efficiency, ultimately contributing to the broader goals of sustainable and intelligent transportation systems \cite{zhu2018modeling}. A key challenge in this field is to replicate human driving styles within established safety parameters, ultimately fostering a seamless integration of these vehicles into the traffic ecosystem. To accomplish this, incorporating driver models that accurately represent individual driving behaviors and trajectories is essential, paving the way for more intuitive and adaptable autonomous systems \cite{he2018human}.

\begin{figure}
\centering
\includegraphics[width=3.5in]{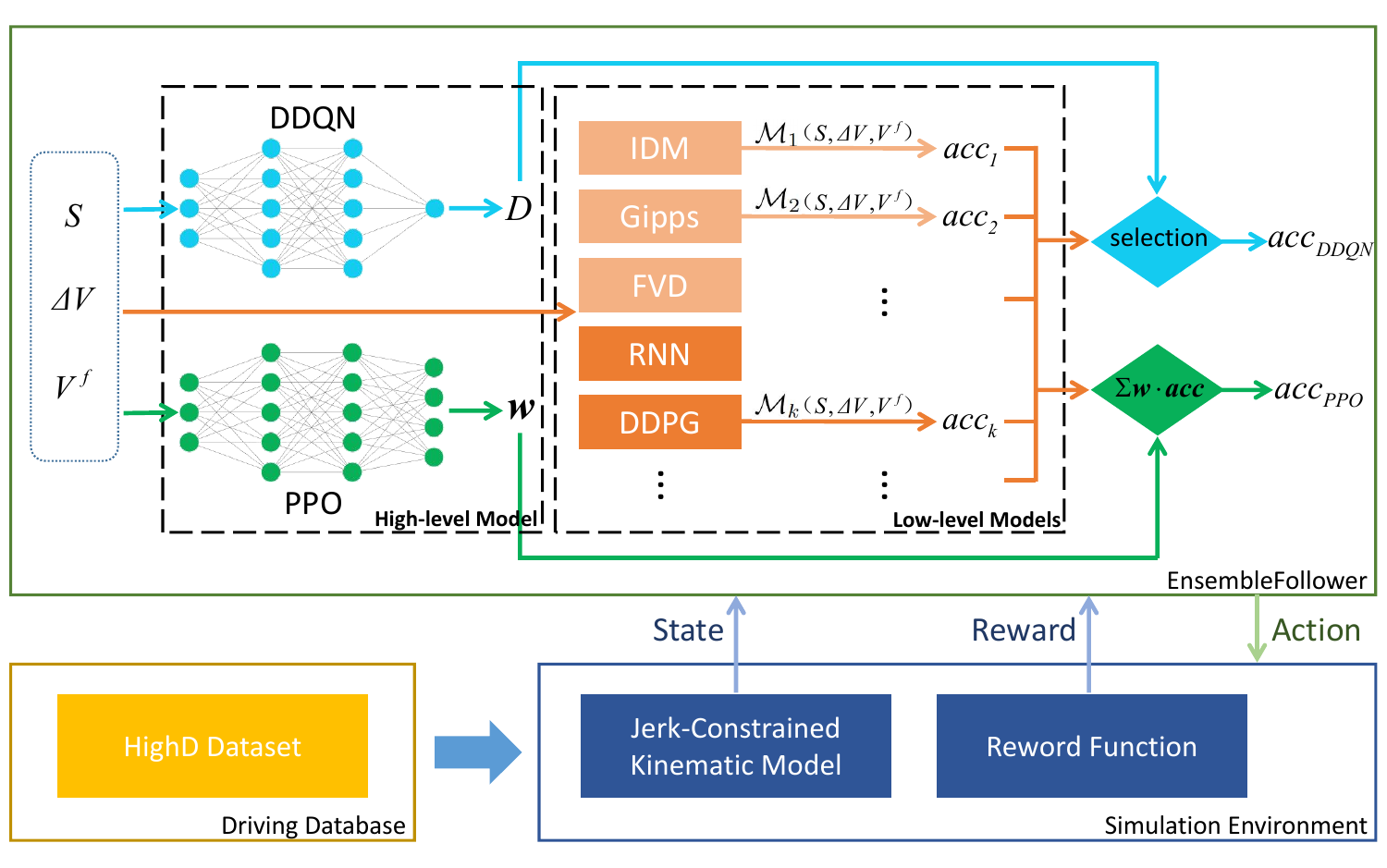}
\caption{The EnsembleFollower framework uses RL algorithms like DDQN or PPO to coordinate between multiple low-level car-following models. Two ways of coordination are proposed: discrete choice (flow in light blue using DDQN) and convex combination (flow in green using PPO). Interactions with a simulator based on HighD dataset is utilized for training the RL agents.}
\label{Figure 1}
\end{figure}

Car-following models have significantly contributed to our understanding of longitudinal vehicle motion and microscopic traffic simulation \cite{brackstone1999car}. However, the representative methods in this domain come with inherent limitations tied to their foundational principles. \emph{Rule-based car-following models, while essential for quick simulations and analytical conclusions, often suffer from limited flexibility and accuracy, stemming from their parsimonious design that cannot fully capture the multifaceted intricacies of the car-following process \cite{wang2017capturing, zhu2018human, chong2011simulation, zhang2023review}.} For instance, the Wiedemann model \cite{Wiedemann_1974}, employed by numerous popular microscopic traffic simulation tools \cite{ag2012vissim}, often exhibits behaviors that diverge from empirical observations \cite{zhu2018modeling}. Similarly, the well-known Velocity Difference Model \cite{jiang2001full} displays considerable sensitivity to its parameter settings, with certain configurations at higher desired speeds possibly resulting in vehicle collisions \cite{kesting2008calibrating}. On the other hand, \emph{data-driven models, built on learning from observed behaviors, grapple with generalization challenges. Although adept at replicating behaviors within their training data, these models falter when confronted with novel or unrepresented scenarios, leading to potential safety concerns and performance inconsistencies \cite{ye2019automated, zhao2022personalized, jurj2021increasing, le2022survey}.} For example, models developed via Imitation Learning \cite{wang2017capturing} circumscribe driving competencies to human comprehension, thereby narrowing the representation to the behaviors of a specific subset of drivers rather than encompassing a broad spectrum of driving skills. As a result, these models often exhibit diminished robustness in scenarios that markedly deviate from their training cases \cite{le2022survey}. Furthermore, Reinforcement Learning (RL) approaches \cite{zhu2018human} are similarly plagued by issues of generalization. While data augmentation techniques \cite{raileanu2021automatic, hansen2021generalization} have been proposed as potential remedies, they merely offer a partial resolution. The inherent dichotomy of driving safety versus traffic efficiency further complicates this generalization challenge \cite{li2018situation}.

Nevertheless, it is essential to note that while no single model can address all of the aforementioned limitations, each model possesses its own strengths and weaknesses depending on the specific car-following scenario. For instance, the widely recognized Intelligent Driver Model (IDM) \cite{treiber2000congested} provides robustness in varying driving conditions, which is a perfect candidate for basic performance guarantee. The Gipps model \cite{gipps1981behavioural} tends to be more effective in preventing collisions due to its safety-distance foundation \cite{zhu2018modeling}, showing the potential to act as a safety lock. Beyond rule-based models, emerging learning-based ones also have their unique advantages, e.g., the Recurrent Neural Network (RNN)-based model is good at simulating hysteresis phenomena, which is credited to its long memory and contributes to an improved accuracy \cite{wang2019long}. Simultaneously, recent research has demonstrated the effectiveness of combining RL and Ensemble Learning (EL) techniques across diverse domains, including finance, Internet of Things, and environmental areas \cite{carta2021multi, jiang2021distributed, Sharma_Singh_Jain_2022}. By integrating multiple models, this method can thoroughly explore the problem space and attain robust performance across diverse scenarios.

On account of the above analysis, we propose an RL-based hierarchical planning framework, EnsembleFollower, as described in Fig. \ref{Figure 1}, to combine the best of both individual car-following models and RL. The high-level model in the framework acts as a coordinator, which chooses the most suitable low-level car-following model to perform an action on the grounds of the current state, or integrates the results of multiple car-following models based on carefully calculated weights. Furthermore, we observed that car-following simulations utilizing data-driven approaches with the conventional kinematic model yielded erratic and unnatural speed profiles. Such anomalous behavior arose predominantly from abrupt accelerations. To enable more scientifically accurate and authentic car-following simulations, we integrated a jerk constraint into the kinematic model.

To our knowledge, this work represents the first attempt to combine RL with multiple car-following models for autonomous driving, and validate it with natural data. The contributions we make in this paper are threefold:

\begin{itemize}
\item We present a novel hierarchical car-following framework named EnsembleFollower that integrates a high-level RL-based decision-maker with an ensemble of hybrid low-level car-following models. This adaptive and efficient planner capitalizes on the strengths of various car-following models, enabling it to make well-informed decisions in different kinds of car-following situations.
\item We propose a jerk-constrained kinematic model to rectify discrepancies in learning-based approaches like RNN and RL, ensuring more scientifically precise and convincing car-following simulations.
\item We compare the EnsembleFollower framework with rule-based methods and data-driven methods on real-world driving data to demonstrate its ability to reproduce accurate human-like driving behavior.
\end{itemize}

The rest of this paper is organized as follows. Section \ref{section:related} provides an overview of related methods. Section \ref{section:formulation} presents the human-like car-following problem and the jerk-constrained kinematic model. Section \ref{section:approach} describes the proposed EnsembleFollower framework. Section \ref{section:expe} introduces the experiment design and analyzes the efficacy of the proposed approach. Section \ref{section:conclusion} draws the conclusions.

\section{Related Work} \label{section:related}
\subsection{Rule-based Car-Following Models}
Car-following models play a critical role in delineating the behavior of a following vehicle (FV) in response to the actions of a leading vehicle (LV)\cite{saifuzzaman2014incorporating}. These models form the foundation of microscopic traffic simulations and serve as vital reference points for intelligent car-following systems \cite{li2012microscopic}.

The investigation of car-following models began in 1953 \cite{pipes1953operational}, and since then, a multitude of models have emerged, including the IDM \cite{treiber2000congested}, the Optimal Velocity Model \cite{bando1995dynamical}, the Gipps model \cite{gipps1981behavioural}, and models proposed by Helly \cite{helly1959simulation} and Wiedemann \cite{Wiedemann_1974}. Recent research has shown improvements and exhibited efficacy \cite{li2018situation, ngoduy2019langevin}. For a comprehensive review and historical overview of car-following models, readers are referred to the works of Zhu \cite{zhu2018modeling}, Li and Sun \cite{li2012microscopic}, as well as Saifuzzaman and Zheng \cite{saifuzzaman2014incorporating}.

\subsection{Data-Driven Approaches}
RNNs have become a pivotal tool in the domain of deep learning, particularly for tasks that involve sequential data or where past information influences current decision-making. They possess the distinctive capability to maintain the memory of previous inputs, making them uniquely suited for modeling temporal dependencies \cite{sherstinsky2020fundamentals}. In the realm of autonomous driving, RNNs have been employed for a multitude of tasks, ranging from \cite{kim2017probabilistic} to fulfilling car-following tasks accurately \cite{wang2017capturing}. These networks can aptly capture the evolving dynamics of traffic, thus aiding in real-time decision-making.

More recently, RL has emerged as a prominent paradigm for intelligent transportation studies, ranging from coordinating the charging recommendation for electric vehicles to learning decision-making policies for autonomous vehicles \cite{sutton1999reinforcement, zhang2021intelligent, kendall2019learning}. RL focuses on how an agent can learn an optimal policy for decision-making through interactions with the environment, aiming to maximize cumulative rewards over time. In the context of autonomous driving, the agent is the vehicle's control system, and the environment encompasses other vehicles, pedestrians, infrastructure, and various road conditions. Most RL-related problems are formulated as Markov Decision Process (MDP), which is, at each discrete time step $t$, the agent perceives its current state $s_t$ and selects an appropriate action $a_t$ from the available action space $A$, guided by a policy $\pi(a_t | s_t)$ that maps states to actions. Upon executing the selected action $a_t$, the system transitions to a subsequent state $s_{t+1}$, and the agent receives a reward signal $r_t$. This iterative process persists until a terminal state is encountered, at which point the agent's episode is reset. The primary objective of the agent is to optimize its policy so as to maximize the accumulated reward $R_t = \sum_{k=0}^{\infty} \gamma^k r_{t+k}$ over time, which is discounted by a factor $\gamma \in (0, 1]$ to account for the relative importance of immediate versus future rewards in the decision-making process \cite{sutton1999reinforcement}.

A prominent algorithm within the value-based RL paradigm is Q-learning \cite{watkins1992q}, which aims to estimate the optimal action-value function iteratively, $Q(s,a)$, representing the expected return for taking action $a$ in state $s$ and following the optimal policy thereafter. The Q-values are updated according to the Bellman Equation \eqref{eqn:1}, and the optimal policy selects the action with the largest $Q(s, a)$ to achieve the maximum expected future rewards \cite{sutton1999reinforcement}.

\begin{equation}
\label{eqn:1}
Q(s, a)=E\left[r+\gamma \max _{a^{\prime}} Q\left(s^{\prime}, a^{\prime}\right)\right].
\end{equation}

Deep Reinforcement Learning (DRL) has evolved into a powerful tool for solving complex planning and decision-making problems in various domains, including autonomous driving. DRL combines the generalization capabilities of deep neural networks (DNNs) with the sequential decision-making framework of RL to efficiently learn optimal control policies \cite{sutton1999reinforcement}. Building on this foundation, techniques like the Double Deep Q-Network (DDQN) have emerged. This method refines the original Deep Q-Network by incorporating the `double' component, astutely addressing Q-learning's overestimation bias and effectively ensuring more stable and performant learning \cite{van2016deep}. 

DDQN's advancements notwithstanding, other methodologies like Deep Deterministic Policy Gradient (DDPG) have tailored solutions for specific challenges. Particularly suited for continuous action spaces, DDPG merges principles from Deep Q-Learning and policy gradients. Through its actor-critic framework, DDPG creates a synergy where the actor provides deterministic state-to-action mappings, while the critic offers a comprehensive action assessment, ensuring adept exploration of action domains \cite{lillicrap2015continuous}.

Amidst these developments, Proximal Policy Optimization (PPO) stands as a testament to the pursuit of stability in policy gradient. By emphasizing controlled policy updates, PPO averts potential pitfalls associated with drastic changes. This approach, which methodically constrains the policy update ratio, marries the ideals of exploration and exploitation, sans the complexities of higher-order optimization \cite{schulman2017proximal}.

These deep variants of RL have become indispensable tools for researchers and engineers in the quest to develop robust and efficient autonomous driving systems. Following Zhu et al.'s application of DDPG to human-like autonomous car-following \cite{zhu2018human}, several studies have investigated applying DRL in car-following directly. Soft Actor-Critic algorithms were employed for speed control using naturalistic driving data \cite{wang2022velocity}, while RL-based longitudinal driver models were tested on a real car to verify efficacy \cite{xie2021modeling}. Additionally, Spatial-temporal Reinforcement Learning was designed to address `phantom' traffic jams in single-lane traffic environments \cite{xu2021patrol}. Alongside human-like driving, alternative objectives for autonomous car-following were introduced via novel reward functions, such as safe, efficient, and comfortable driving \cite{zhu2020safe, yavas2023toward}. Therefore, RL is an appropriate candidate for high-level policy in autonomous driving, as it can learn from experience to identify the most suitable low-level model when receiving the environment state.

\subsection{Hierarchical Planning and Ensemble Learning}
Hierarchical planning that combines DRL with other algorithms has become another ongoing direction for tackling diverse challenges in autonomous driving. For instance, Hierarchical Reinforcement Learning was used to improve learning efficiency by sharing network architecture and weights among low-level models \cite{qiao2020hierarchical, nosrati2018towards}. Hierarchical Program-triggered Reinforcement Learning, proposed in \cite{gangopadhyay2021hierarchical}, can manage relatively simple tasks with multiple low-level RL agents. In \cite{wang2020learning}, a high-level RL policy was integrated with a sampling-based motion planner for solving long-horizon issues. Similarly, the study in \cite{zhang2021multi} combined the $A^*$ algorithm and RL algorithms to address multi-task, long-range driving problems. In \cite{peng2022integrated}, a high-level Dueling Double Deep Q-Network (D3QN) agent responsible for lane-changing decisions was connected to low-level DDPG agents in charge of speed control. Furthermore, a hierarchical planning framework was assessed in intersection and roundabout scenarios to demonstrate safety \cite{li2021safe}. The utilization of hierarchical planning in autonomous driving enables the combination of high-level decision-making with low-level performers, leading to improved performance and adaptability in various driving situations.

EL refers to a technique in which multiple models, often called `base learners', are trained to solve the same problem and are subsequently combined to improve overall performance. The primary idea is that by aggregating diverse models, the ensemble can capitalize on the strengths and mitigate the weaknesses of individual learners. This approach was systematically introduced by \cite{Krogh_Vedelsby_1994}, with significant advancements made by Random Forests \cite{breiman2001random}. Inspired by the idea of EL, we intend to construct a novel hierarchical planning framework that operates effectively in diverse car-following conditions by employing the fundamental concepts of assembling hybrid low-level models.

\section{Problem Formulation}\label{section:formulation}

The importance of achieving human-like driving behavior in autonomous vehicles stems from the need to maintain consistency and safety on the road. Human-like driving not only offers passengers a comfortable and reliable autonomous experience but also facilitates more predictable interactions between human drivers and autonomous vehicles, allowing for natural and harmonious coexistence on the roads. In this paper, we focus on the human-like driving problem in various car-following conditions. 

\subsection{Conventional Kinematic Model For Car-Following}
We formulate the problem as an MDP as mentioned in the related work section. The state $s_t$ of the car-following environment at a given time step $t$ is characterized by the speed $V_t^f$ of the FV, and the spacing $S_t$ and relative speed $\Delta V_t$ between the LV and the FV. The action $a_t$ corresponds to the longitudinal acceleration $acc_t^f$ of the FV. Utilizing the current state and action at a particular time step, the subsequent state can be updated through a conventional discrete-time kinematic model as described in Equation \eqref{eqn:2}:

\begin{equation}
\label{eqn:2}
\begin{aligned}
& V_{t+1}^f=V_t^f+acc_t^f \cdot \Delta T \\
& \Delta V_{t+1}=V_{t+1}^l-V_{t+1}^f \\
& S_{t+1}=S_t+\frac{\Delta V_t+\Delta V_{t+1}}{2} \cdot \Delta T,
\end{aligned}
\end{equation}

\noindent where the LV speed $V^l$ is provided by the empirical dataset. In this study, we set $\Delta T$, the simulation time interval, to 0.04 seconds according to the HighD dataset.

The reward function $r(s, a)$ in MDP assigns a scalar value reflecting the instant gain of a transition from the initial state $s_t$ to a subsequent state $s_{t+1}$ caused by the action $a_t$. We design the reward function in this study to encourage human-like driving by minimizing the discrepancy between the values of observed and simulated behavior, as detailed in Equation \eqref{eqn:3}, where $S_t^{obs}$ and $V_t^{obs}$ give the observed spacing and FV speed. The overall objective is to maximize the expected discount reward $R = \sum_{t=0}^{\infty} \gamma^t r_t$, where $\gamma \in (0,1]$ is the discount factor in the MDP problem, and derive the optimal policy according to expected reward.

\begin{equation}
\label{eqn:3}
\begin{cases}
r_t=-\log \left(\left|\frac{S_t-S_t^{obs}}{S_t^{obs}}\right|\right)\text{, by spacing discrepancy} \\
r_t=-\log \left(\left|\frac{V_t^f-V_t^{obs}}{V_t^{obs}}\right|\right)\text{, by speed discrepancy.}
\end{cases}
\end{equation}

To assess the performance of each car-following model, we follow the approach in \cite{zhu2018human} and compare the simulated and observed spacings $S$ and FV speeds $V_f$ defined above. We employ the Root Mean Square Percentage Error (RMSPE) of spacing and FV speed, given by Equation \eqref{eqn:4}, as our evaluation metric. Similar to Equation \eqref{eqn:3}, $S_t^{sim}$ and $V_t^{sim}$ represent the simulated spacing and FV speed at time step $t$, and $N$ denotes the total number of simulation steps. Additionally, collision rate is also calculated to measure the safety of the model, which is defined as the number of car-following events with spacing between vehicles becoming negative divided by the total number of car-following events in the test dataset. These metrics provide a comprehensive assessment of the model's performance and ensure the validity of the research findings.

\begin{equation}
\label{eqn:4}
\begin{aligned}
& \text { RMSPE of spacing }=\sqrt{\frac{\sum_{t=1}^N\left(S_t^{s i m}-S_t^{o b s}\right)^2}{\sum_{t=1}^N\left(S_t^{o b s}\right)^2}} \\
& \text { RMSPE of speed }=\sqrt{\frac{\sum_{t=1}^N\left(V_t^{s i m}-V_t^{o b s}\right)^2}{\sum_{t=1}^N\left(V_t^{o b s}\right)^2}}.
\end{aligned}
\end{equation}

\begin{figure}
\centering
\includegraphics[width=3.4in]{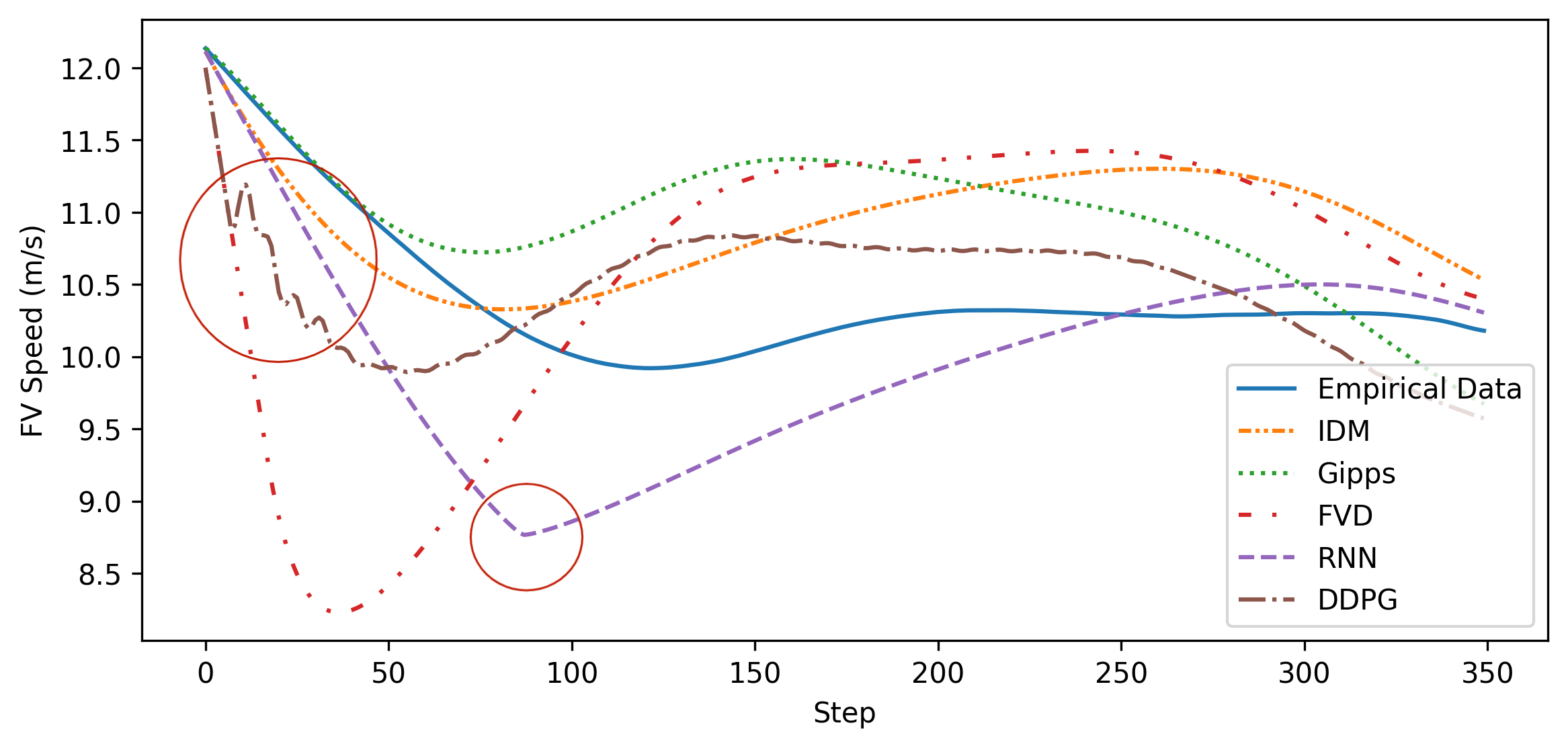}
\caption{Unnatural speed profile observed in the conventional kinematic model.}
\label{unnatural_speed}
\end{figure}

\begin{figure}
\centering
\includegraphics[width=3.4in]{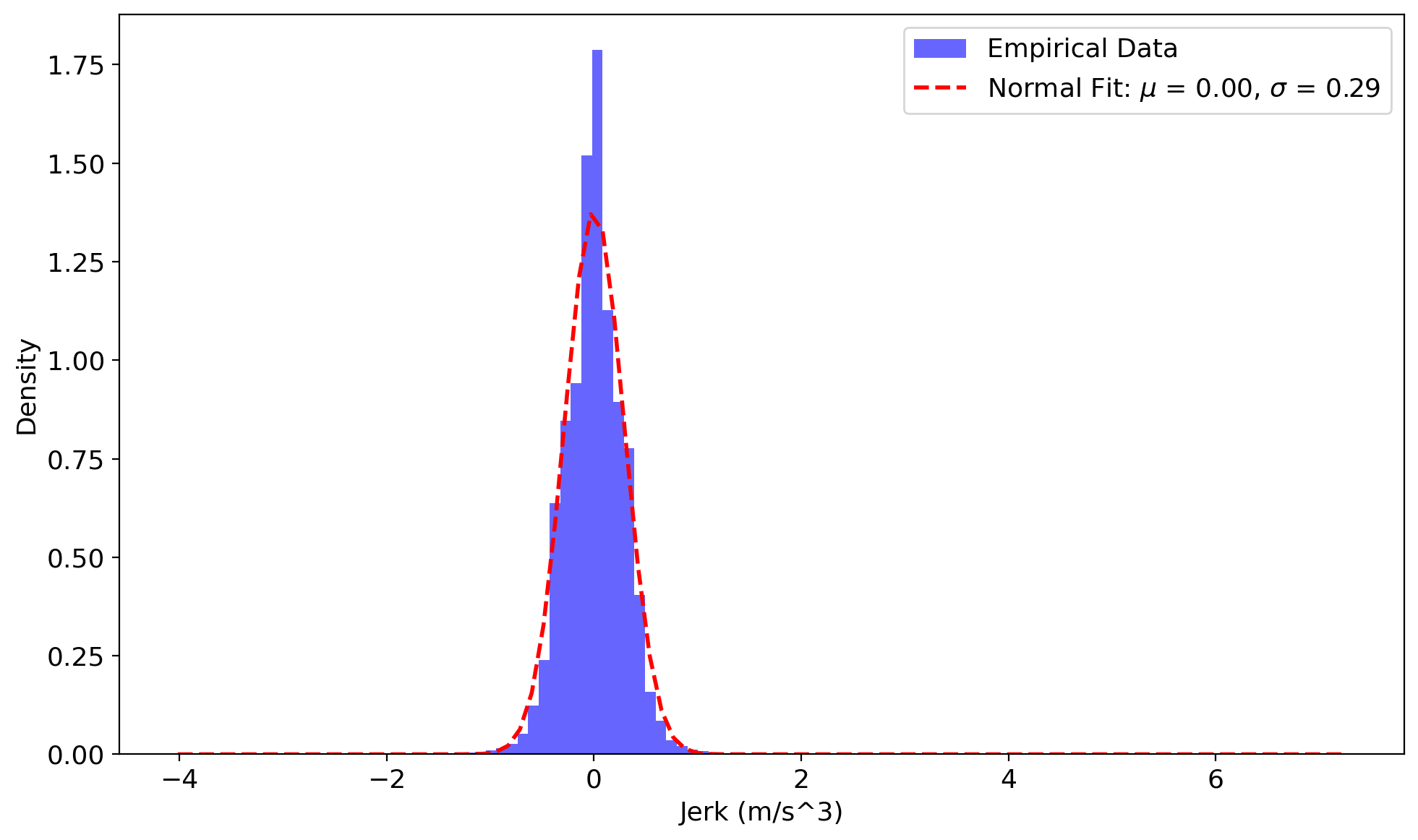}
\caption{Probability density distribution of the jerk in car-following events extracted from the HighD dataset mirrors a normal curve.}
\label{jerk_dist}
\end{figure}

\subsection{Adding the Constraint of Jerk}
When implementing the conventional kinematic model, notable discrepancies were observed in the outcomes produced by data-driven approaches such as RNN and DDPG. Specifically, these methods sometimes generated irregular and unrealistic speed profiles during certain car-following scenarios, as depicted in Fig. \ref{unnatural_speed}. It is worth mentioning that such anomalies are seldom observed with rule-based car-following models. The primary reason is that the rule-based models tend to produce accelerations adhering to Lipschitz continuity, resulting in smooth speed transitions. On the contrary, learning-based models occasionally generate accelerations at each timestep that starkly deviate from their preceding values. Such behavior results in a high jerk – the rate of change of acceleration. This not only breaches fundamental physical constraints but also detracts from the realism of the driving simulation, leading to uncharacteristic speed profiles.

In the current experimental setup, where an acceleration range of [-4.0, 4.0]\,m/s$^{2}$ and a simulation frequency of 25\,Hz are adopted, the theoretical jerk can surge to an extreme value of 200\,m/s$^{3}$ without any imposed limitations. However, a closer examination of car-following events, extracted from the HighD dataset and illustrated in Fig. \ref{jerk_dist}, reveals that the jerk distribution closely mirrors a normal curve, bounded within the range of [-3.50, 6.74]\,m/s$^{3}$. It's essential to recognize that the maximum value of jerk depends on the vehicle systems and the operation conditions \cite{hoberock1976survey, lin2019longitudinal}. Some recent works have explored the combination of acceleration command dynamics and DRL for vehicle control \cite{buechel2018deep, wei2018design}. However, a time constant in the acceleration dynamics equation needs to be determined based on each vehicle's powertrain system. Given these observations, we propose a concise but effective kinematic model, which integrates the model with a jerk constraint (Equation \eqref{eqn:jerk-b}) to ensure a more scientifically precise and genuine simulation:

\begin{subequations}
    \begin{align}
    & {jerk_t^f}=\frac{acc_t^f-acc_{t-1}^{CLIP}}{\Delta T} \label{eqn:jerk-a} \\
    & {jerk_t^{CLIP}}=\text{clip}(jerk_t^f, jerk_{min}, jerk_{max}) \label{eqn:jerk-b} \\
    & {acc_t^{CLIP}}=acc_{t-1}^{CLIP}+{jerk_t^{CLIP}} \cdot \Delta T \label{eqn:jerk-c} \\
    & V_{t+1}^f=V_t^f+acc_t^{CLIP} \cdot \Delta T \label{eqn:jerk-d} \\
    & \Delta V_{t+1}=V_{t+1}^l-V_{t+1}^f \label{eqn:jerk-e} \\
    & S_{t+1}=S_t+\frac{\Delta V_t+\Delta V_{t+1}}{2} \cdot \Delta T \label{eqn:jerk-f},
    \end{align}
\end{subequations}

\noindent where $acc_t^f$ is the acceleration produced by car-following model, $acc_0^{CLIP}$ equals $acc_0^f$ as an edge case, $jerk_{min}$ and $jerk_{max}$ are constants and set to -10 and 10 respectively based on the jerk distribution and without loss of generality. This jerk-constrained kinematic model is imperative, regardless of any potential jerk-associated loss or reward functions that might be incorporated by learning-based algorithms.

\section{Proposed Approach}\label{section:approach}
\subsection{Hierarchical Framework}
This paper presents EnsembleFollower, an RL-based hierarchical behavior planning framework, to address the human-like car-following problem, as depicted in Fig. \ref{Figure 1}. The simulation environment is based on the proposed jerk-constrained kinematic model and backed by a driving database extracted from the HighD dataset. The EnsembleFollower framework receives the current environmental state as input and a reward as instant feedback, and then generates actions for the FV to execute. As described in the last section, the input $s_t$ contains the spacing $S$, the relative speed $\Delta V$, the FV speed $V^f$, and the action output $a_t$ is the FV longitudinal acceleration. This hierarchical framework comprises an ensemble of low-level car-following models and a high-level RL agent responsible for managing them. Two styles of management, including discrete choice and convex combination, are proposed and experimented, which will be explained in detail in the following subsections. Our objective of training the RL agent is to identify a policy $\pi^*$ as described in Equation \eqref{eqn:policy}, where $r$ is the reward function introduced in the last section. The desired policy is capable of assessing the current environmental conditions and judiciously leverages the most appropriate models at each step for autonomous car-following.

\begin{equation}
\label{eqn:policy}
\pi^*=\arg \max _\pi \mathbb{E}\left[\sum_{t=0}^{\infty} \gamma^t r\left(s_t, \pi\left(s_t\right)\right)\right].
\end{equation}

\subsection{Low-Level Models}
In the proposed EnsembleFollower framework, the ensemble of low-level models is composed of $k$ car-following models, symbolized as $\{\mathcal{M}_1, \mathcal{M}_2, ..., \mathcal{M}_k\}$. These models function as specialized agents within the hierarchical structure, handling distinct circumstances delegated by the high-level coordinator. The low-level models receive a state input $s_t$ and return a corresponding action $a_t$, which is the longitudinal acceleration in the car-following case, at a given time step $t$.

To showcase the efficacy of our hierarchical framework, we have incorporated five representative car-following models, including both rule-based models and data-driven models. The rule-based models were the IDM \cite{treiber2000congested}, Gipps' model \cite{gipps1981behavioural}, and the Full Velocity Difference (FVD) model. The FVD model shares similarities with the Optimal Velocity Model introduced in \cite{bando1995dynamical}. The data-driven models consisted of a RNN model \cite{wang2017capturing} and a DDPG model \cite{zhu2018human}. Prior to training the high-level RL agent, it is essential to calibrate or train the configurable parameters of these low-level models, for the purpose of enhancing the stability of the high-level agent training process. Further elaboration of calibration and training is provided in experiments section.

\begin{algorithm}[t]
\label{alg:1}
\caption{EF-DDQN Training Algorithm}\label{algorithm}
\KwInit{the low-level models $\mathcal{M}_1,\mathcal{M}_2, ..., \mathcal{M}_k$; the initial policy containing the Q-networks with weights $\theta$ and $\theta'$, set $\theta' \leftarrow \theta$.}
\KwInst{a replay buffer $\mathcal{D}$ with a capacity of $l_D$.}
\For {episode $\leftarrow 1$ \KwTo $E$}{Reset the simulation environment $env$ and get an initial car-following state $s_1$: initial gap, follower speed, and relative speed. \\
\For{step $t \leftarrow 1$ \KwTo $T$}{Select a model number $D_t \leftarrow \arg\max_{D_t}Q_\theta(s_t, D_t)$ according to the $\epsilon - greedy$.\\
With $s_t$, the selected model $\mathcal{M}_{D_t}$ outputs follower acceleration $acc_t$. \\
Apply acceleration $acc_t$ and update to new state $s_{t+1}$ based on jerk-constrained kinematic model in $env$. \\
Calculate reward $r_t$ based on the disparity between observed ground truth and new state $s_{t+1}$ in $env$. \\
Save the transition $[s_t \quad D_t \quad r_t \quad s_{t+1}]$ into $\mathcal{D}$.\\
Sample a mini-batch with size $P$ of transitions $[s_i \quad D_i \quad r_i \quad s_{i+1}]$ from $\mathcal{D}$. \\
Update $\theta$ by minimizing loss: $L \leftarrow \frac{1}{P} \sum_i\left(y_i-Q_{\theta}\left(s_i, D_i\right)\right)^2$, where $y_i \leftarrow r_i+\gamma Q_{\theta^\prime}\left(s_{i+1}, \arg\max_{D}Q_{\theta}(s_{i+1}, D)\right)$ \\
\If{$t$ mod $target \_ update \_ frequecy == 0$}{$\theta' \leftarrow \theta.copy()$}}
Record the cumulative reward in this episode.\\}
\end{algorithm}

\subsection{High-Level RL Algorithm}
The core of this framework is a high-level RL algorithm that orchestrates the coordination between hybrid low-level models. Specifically, two separate approaches are proposed for effective integration.

\subsubsection{Discrete Choice}
The RL algorithm processes the environmental state and generates a decision $D$ for selecting the most appropriate low-level model at the moment. As a result, the action space of the RL agent, also known as the domain of $D$, becomes a set of discrete model identifiers, e.g., $\{1, 2, \dots, k\}$, where each identifier corresponds to a unique low-level model. Within the hierarchical structure, the actions produced by the RL agent are considered intermediate, while the longitudinal accelerations that stem from the chosen low-level models work directly on the jerk-constrained environment, as shown in Equation \eqref{eqn:acc_ddqn}:

\begin{equation}
\label{eqn:acc_ddqn}
acc_t^f = acc_t^D, D \in \{1, 2, ..., k\}.
\end{equation}

\subsubsection{Convex Combination}
The RL agent assigns weights for each low-level model based on the environmental state, where the weights are non-negative scalars and sum up to one. After that, the output of the EnsembleFollower framework $acc_t^f$ is calculated based on Equation \eqref{eqn:acc_ppo}:

\begin{equation}
\label{eqn:acc_ppo}
\begin{aligned}
& acc_t^f = \sum_{i=1}^{k}w_t^i \cdot acc_t^i, \\
& \text{s.t. } \sum_{i=1}^{k}w_t^i = 1, w_t^i \geq 0, i = 1, 2, ..., k, \\
\end{aligned}
\end{equation}

\noindent where $w_t^i$ is the weight assigned for the $i$th model and $acc_t^i$ is the acceleration generated by this model at time $t$.

In general, a variety of model-free RL algorithms can be applied for training the high-level agent. However, we choose the DDQN algorithm for the proposed discrete choice approach as it is suitable for the problems with discrete action space and offers reduced variance. In terms of the convex combination method, the PPO algorithm is adopted due to its increased stability in policy gradient. The EnsembleFollower agents trained with DDQN and PPO are called EF-DDQN and EF-PPO respectively in the following sections. The training procedure for EF-DDQN is outlined in Algorithm \ref{alg:1}. EF-PPO shares a similar training procedure except that it collects the training cases in an on-policy manner and updates the networks by maximizing $L_t(\theta)$ according to Equation \eqref{eqn:ppo-1}, which is defined as \cite{schulman2017proximal}:

\begin{small}
\begin{equation}
\label{eqn:ppo-1}
\begin{aligned}
& L_t(\theta)=\hat{\mathbb{E}}_t\left[L_t^{C L I P}(\theta)-c_1 L_t^{V F}(\theta)+c_2 B\left(s_t|\theta\right)\right] \\
& L_t^{C L I P}(\theta)=\min \left(p_t(\theta) \hat{A}_t, \operatorname{clip}\left(p_t(\theta), 1-\epsilon, 1+\epsilon\right) \hat{A}_t\right) \\
& \hat{A}_t=\delta_t+(\gamma \lambda) \delta_{t+1}+\cdots+\cdots+(\gamma \lambda)^{T-t+1} \delta_{T-1} \\
& \delta_t=r_t+\gamma V\left(s_{t+1}\right)-V\left(s_t)\right),
\end{aligned}
\end{equation}
\end{small}

\noindent where $V(s)$ is state-value function, $c_1$, $c_2$ are coefficients, $B$ is an entropy bonus, $L_t^{V F}$ denotes a squared-error loss $(V_\theta(s_t)-V_t^{targ})^2$, $p(\theta)$ is the probability ratio $\pi_\theta(a_t|s_t)/\pi_{\theta_{old}}(a_t|s_t)$, $\epsilon$ is a hyperparameter, $\gamma$ is the discount factor, and $\lambda$ is the factor for Generalized Advantage Estimation.

\section{Experiments and Results} \label{section:expe}

\subsection{Experiment Design}
\begin{figure}
\centering
\includegraphics[width=3.4in]{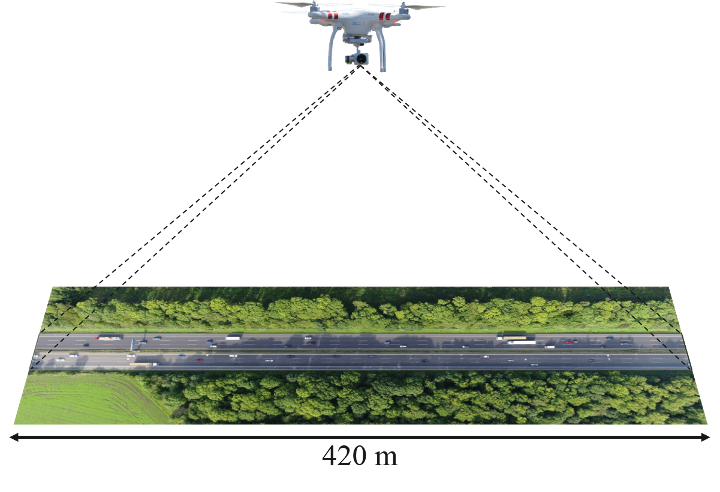}
\caption{HighD dataset is used for training and evaluation.}
\label{Figure 2}
\end{figure}

To evaluate the performance of the proposed EnsembleFollower framework, we utilized the vehicle trajectory data in the HighD dataset \cite{highDdataset}, which is a comprehensive driving dataset released by the Institute of Automotive Engineering at RWTH Aachen University in Germany. It features high-precision information on vehicle positions and speeds, captured through bird's-eye view videos of six different roads around Cologne using a high-resolution 4K camera mounted on an aerial drone, as shown in Fig. \ref{Figure 2}. Advanced computer vision techniques ensure positional inaccuracy is typically less than 10 cm, with Bayesian smoothing used to eliminate noise and smooth motion data. The dataset contains more than 110,500 vehicles captured from six distinct locations and provides automatic extraction of vehicle trajectory, size, type, and maneuvers. While developed primarily for highly automated car safety validation, it is also useful for applications such as traffic pattern analysis and driver model parameterization and is widely used by researchers and automakers to develop self-driving technologies and improve autonomous vehicle safety.

To extract car-following events for the driving database, a car-following filter was applied. This filter identified events where LVs and FVs remained in the same lane for at least 15 seconds, ensuring the car-following persisted long enough for analysis. In addition, to ensure that the recorded events were representative and comparable, those with low speed or stoppage that lasted longer than 5 seconds were filtered. A total of 12,541 car-following events with a frequency of 25\,Hz provided by \cite{chen2023follownet} are utilized for this study.

Based on the driving database, we developed a simulator to interact with the car-following models according to the jerk-constrained kinematic model. When a car-following event is completed in the simulation, we reset the environmental state with another event in the database. The HighD dataset is divided into a training, validation, and testing set, with 70\%, 15\%, and 15\% of the data, respectively. To establish the low-level models, we calibrated or trained them with collision checking on the training set. After that, we trained our RL-based hierarchical models based on the training set and the established low-level models. Next, we evaluated the model candidates on the validation set to determine the best-performing model. Finally, we compared the performance of our RL-based models with the representative car-following models based on the RMSPEs on the testing set.

Given that the RL model's effectiveness may vary according to the chosen reward function, we test both spacing and FV speed for reward. Our results indicate that the model employing speed discrepancy as the reward function demonstrates superior performance, and we adopt this configuration in the following sections.

\subsection{Calibration and Training of Low-Level Models}
In this study, we calibrated several traditional car-following models using the RMSPE of spacing as the objective function, as suggested in previous literature \cite{zhu2018modeling}. The objective function aims to minimize the difference between the observed and simulated spacing.

Table \ref{low-calib} (Appendix A) provides an overview of the behavioral parameters of the calibrated models, including the bounds of these parameters. The parameters include the desired headway, time constant, sensitivity coefficient, and reaction time, among others. To find the optimum values of the model parameters, we implemented a Genetic Algorithm (GA), which searches the parameter space to identify the parameter values that minimize the objective function. The GA parameters for calibration include a population size of 100, a maximum generation number of 100, a stall generation number of 100 and a mutation probability of 0.2. Notably, a crash penalty was added to the objective function to avoid undesirable solutions leading to collisions.

The hyperparameters of training the RNN model and the DDPG model are listed in Table \ref{low-train} (Appendix A). 25 most recent states, corresponding to a reaction time of 1 second, were fed into the neural networks. The RNN model was trained with a network containing one Long Short-Term Memory layer, converged in 20 training epochs. The DDPG model achieved its optimal performance after 1,000,000 training steps with setting gamma to 0.96.

\begin{table*}
\centering
\caption{\label{table-2}Hyperparameters and Corresponding Descriptions for EF-DDQN and EF-PPO training.}
\begin{tabular}{llll}
\hline
Model & Hyperparameter & Description & Value \\ \hline
EF-DDQN & Learning rate & Learning rate used by the optimizer & 0.0003 \\
 & Discount factor & Discount factor gamma for estimating Q value & 0.99 \\
 & Minibatch size & Number of training cases used by gradient descent & 4,096 \\
 & Training start & Number of training cases to collect before learning starts & 200,000 \\
 & Replay buffer size & Maximum number of training cases in replay buffer & 1,000,000 \\
 & First hidden layer size & Number of neurons in the first hidden layer & 64 \\
 & Second hidden layer size & Number of neurons in the second hidden layer & 32 \\
 & Train frequency & Frequency (steps) for updating the main network & 4 \\
 & Target update interval & Frequency (steps) for updating the target network & 250 \\
 & Final exploration probability & Final value of random action probability & 0.25 \\ \hline
EF-PPO & Actor learning rate & Learning rate used by the actor optimizer & 0.001 \\
 & Critic learning rate & Learning rate used by the critic optimizer & 0.001 \\
 & Learning rate decay & Whether learning rate decay technique is applied & True \\
 & Discount factor & Discount factor gamma for estimating Q value & 0.99 \\
 & GAE lambda & Parameter for Generalized Advantage Estimation & 0.95 \\
 & Step per collect & Number of training cases to collect before the network update & 5,000 \\
 & Repeat per collect & Number of repeat time for policy learning & 4 \\
 & Minibatch size & Number of training cases used by gradient descent & 2,500 \\
 & First hidden layer size & Number of neurons in the first hidden layer & 64 \\
 & Second hidden layer size & Number of neurons in the second hidden layer & 32 \\
 & Epsilon & Parameter for policy gradient clipping & 0.2 \\
 & Value function coefficient & Weight for value function when calculating loss & 0.25 \\
 & Entropy coefficient & Weight for entropy when calculating loss & 0.01 \\ \hline
\end{tabular}
\end{table*}

\begin{figure}
\centering
\includegraphics[width=3.4in]{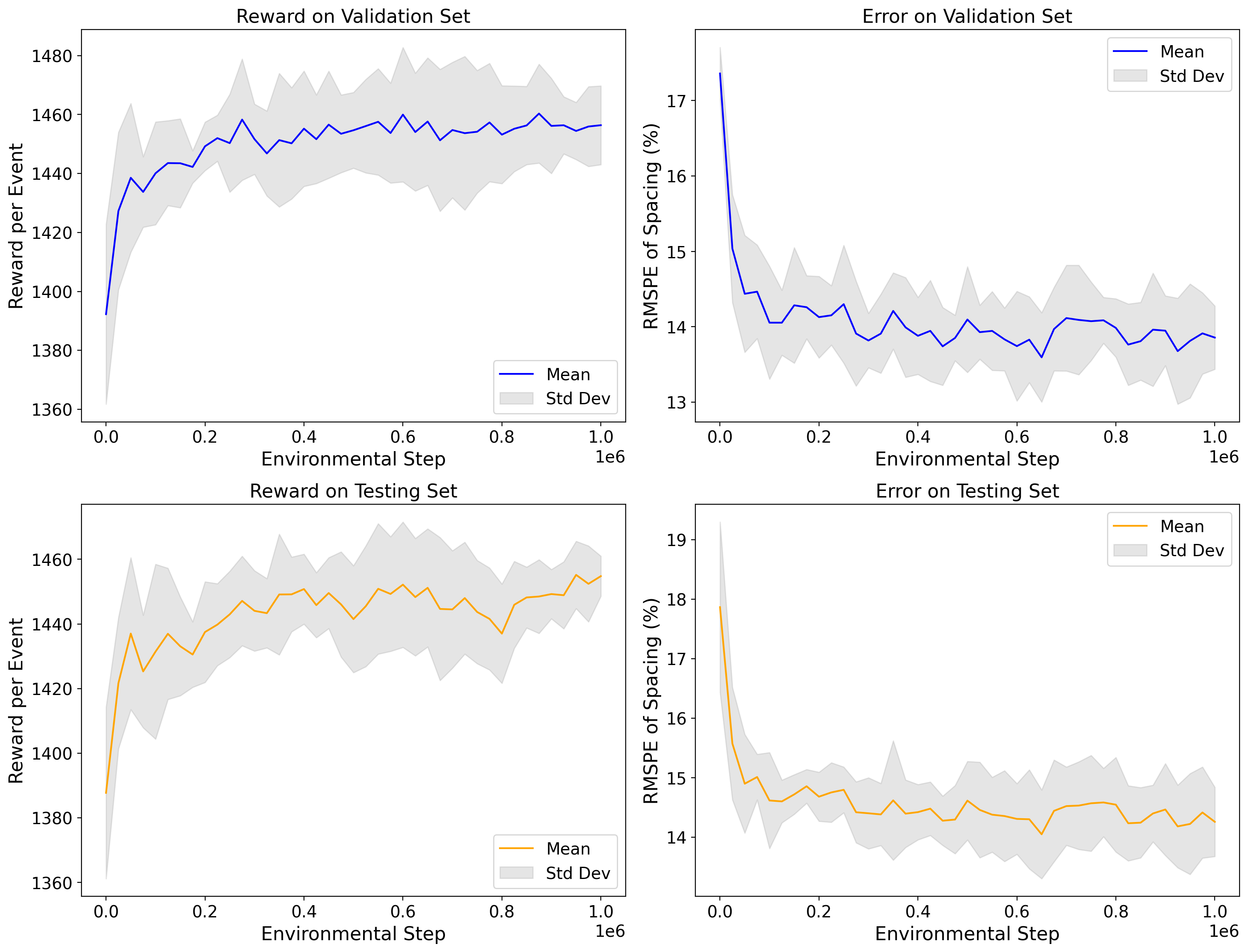}
\caption{Reward and error on validation set and testing set during EF-DDQN training.}
\label{reward_error_dqn}
\end{figure}

\begin{figure}
\centering
\includegraphics[width=3.4in]{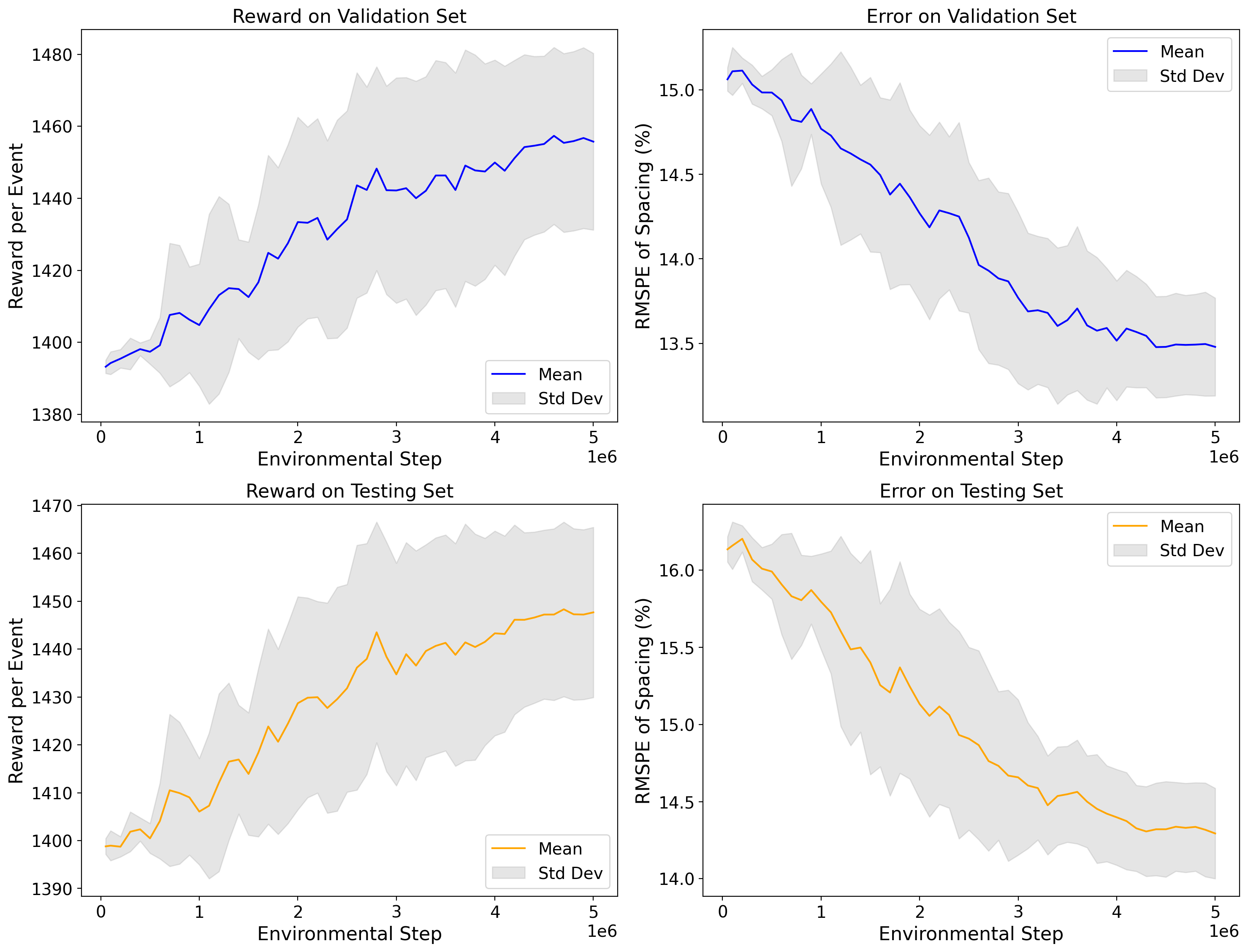}
\caption{Reward and error on validation set and testing set during EF-PPO training.}
\label{reward_error_ppo}
\end{figure}

\subsection{Training of High-Level RL Models}
EF-DDQN utilizes a main network and a target network, which share the same architecture comprising four linear layers: an input layer, an output layer, and two hidden layers. We initially attempted with network architectures containing only one hidden layer, but our subsequent analysis revealed that using two hidden layers significantly improved performance. The input to each neural network consists of environmental states described in the problem definition section, and we adopted a reaction time of 1 second in line with Zhu et al. \cite{zhu2018human}, which indicated that 25 most recent observed states were employed by the framework. The output of EF-DDQN is an integer representing the selected low-level model.

The input of EF-PPO is the same as that of EF-DDQN while the output of EF-PPO is a series of scalars with the length equaling the number of low-level models. The actor network and the critic network in EF-PPO share a similar architecture with the networks in EF-DDQN.

The high-level models' hyperparameter values were determined based on the model's performance on the validation set and presented in Table \ref{table-2}. After that, the best-performing model is compared with other models in the next subsection. Fig. \ref{reward_error_dqn} and Fig. \ref{reward_error_ppo} depict the average event reward and error on both the validation and testing sets during training. Five random seeds were tested for each algorithm and the standard deviation is shown as the shaded area. It can be seen that the average event reward received by EF-DDQN agent increased rapidly and converged to near-optimal in the early training stage. In contrast, EF-PPO did not exhibit such stable learning, likely due to the dramatically expanded action space compared to the discrete action choices. The larger standard deviations for EF-PPO also suggest a more turbulent learning process. To improve training stability for EF-PPO, a linear learning rate decay technique was implemented based on total environmental steps. Despite signs of convergence for EF-PPO in later training stage with the decay technique, substantially more training steps and much longer training times were required to attain strong performance compared to EF-DDQN. Overall, the plots suggest a stable learning process and reliable generalization ability to unseen situations of the proposed EnsembleFollower framework.

\begin{figure}
\centering
\includegraphics[width=3.4in]{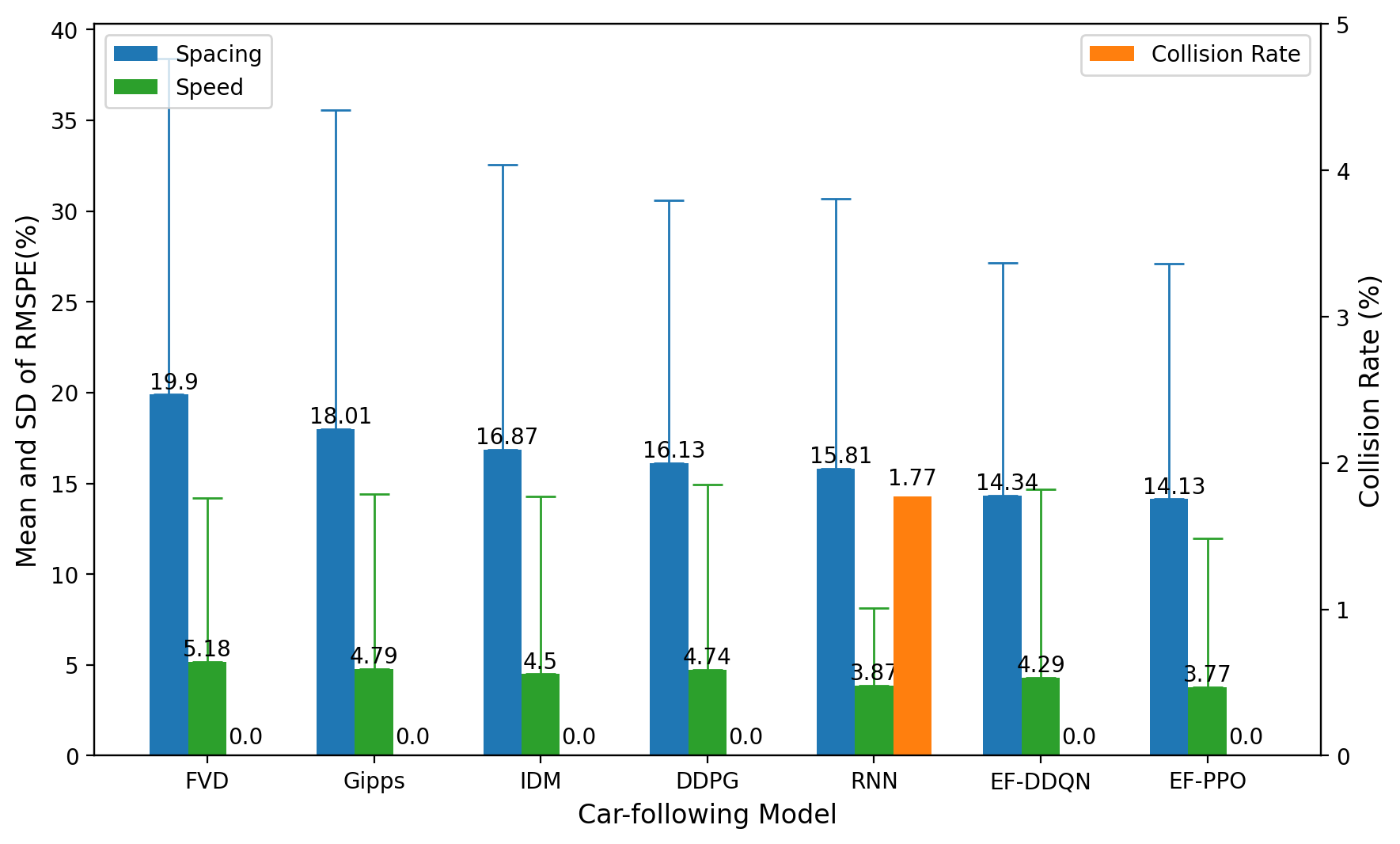}
\caption{Comparison of model performances on the testing dataset.}
\label{model comparison}
\end{figure}

\subsection{Comparison of Model Performances}
In order to evaluate the accuracy of the proposed EnsembleFollower framework in replicating human-like driving behaviors, we conducted a comparison of the RMSPE of spacing, the RMSPE of speed, and the collision rate, on a reserved testing set. Fig. \ref{model comparison} shows the mean and standard deviation of the RMSPEs for EF-DDQN and EF-PPO, as well as three well-established car-following models and two cutting-edge data-driven approaches\cite{wang2017capturing, zhu2018human}.

Our findings indicate that the proposed framework outperformed all investigated models, as proved by its ability to achieve the lowest mean and standard deviation values for the RMSPE of spacing. While the RNN model achieved the lowest RMSPE of speed, it came at the cost of a high collision rate. The EF-PPO model behaved more accurately compared to the EF-DDQN model, especially in terms of the RMSPE of speed, indicating that weighted integration was a more effective technique than discrete categorical options in spite of more effort and cost would be needed during the training process. These results highlight the efficacy of EnsembleFollower in improving the accuracy of car-following models for human-like autonomous driving applications while maintaining zero collision.

\begin{table*}
\centering
\caption{Summary of the EF-DDQN agent’s selections on the testing dataset.}
\begin{tabular}{lllllllll}
\hline
Model & Ratio (\%) & Spacing & LV Speed & Relative Speed & LV Speed Change (\%) \\ \hline
RNN & 32.12 & 38.01 ($\pm$ 15.21) & 20.30 ($\pm$ 5.15) & -0.16 ($\pm$ 0.92) & -37.36 ($\pm$ 35.07) \\
Gipps & 31.08 & 24.92 ($\pm$ 6.52) & 22.80 ($\pm$ 2.28) & -0.35 ($\pm$ 0.64) & -3.12 ($\pm$ 25.57) \\ 
IDM & 15.96 & 23.05 ($\pm$ 12.31) & 18.28 ($\pm$ 6.03) & 0.10 ($\pm$ 0.83) & -5.14 ($\pm$ 49.60) \\
DDPG & 13.82 & 20.16 ($\pm$ 9.12) & 18.75 ($\pm$ 6.54) & -0.04 ($\pm$ 0.57) & 4.84 ($\pm$ 42.38) \\
FVD & 7.02 & 25.85 ($\pm$ 12.29) & 21.05 ($\pm$ 5.23) & -0.02 ($\pm$ 0.74) & -6.72 ($\pm$ 35.43) \\ \hline
\end{tabular}
\label{table:summary}
\end{table*}

\begin{figure}
\centering
\includegraphics[width=3.4in]{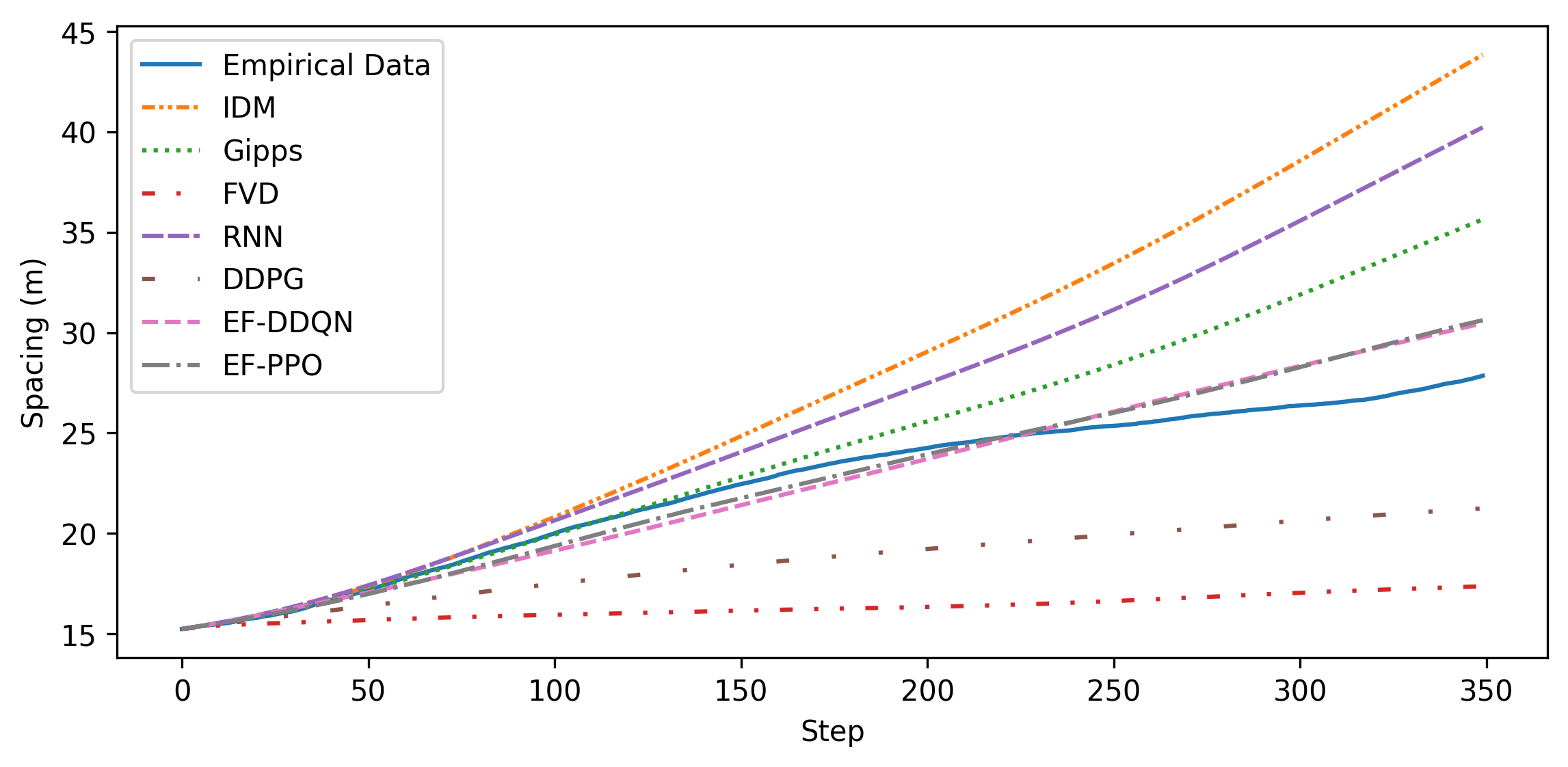}
\includegraphics[width=3.4in]{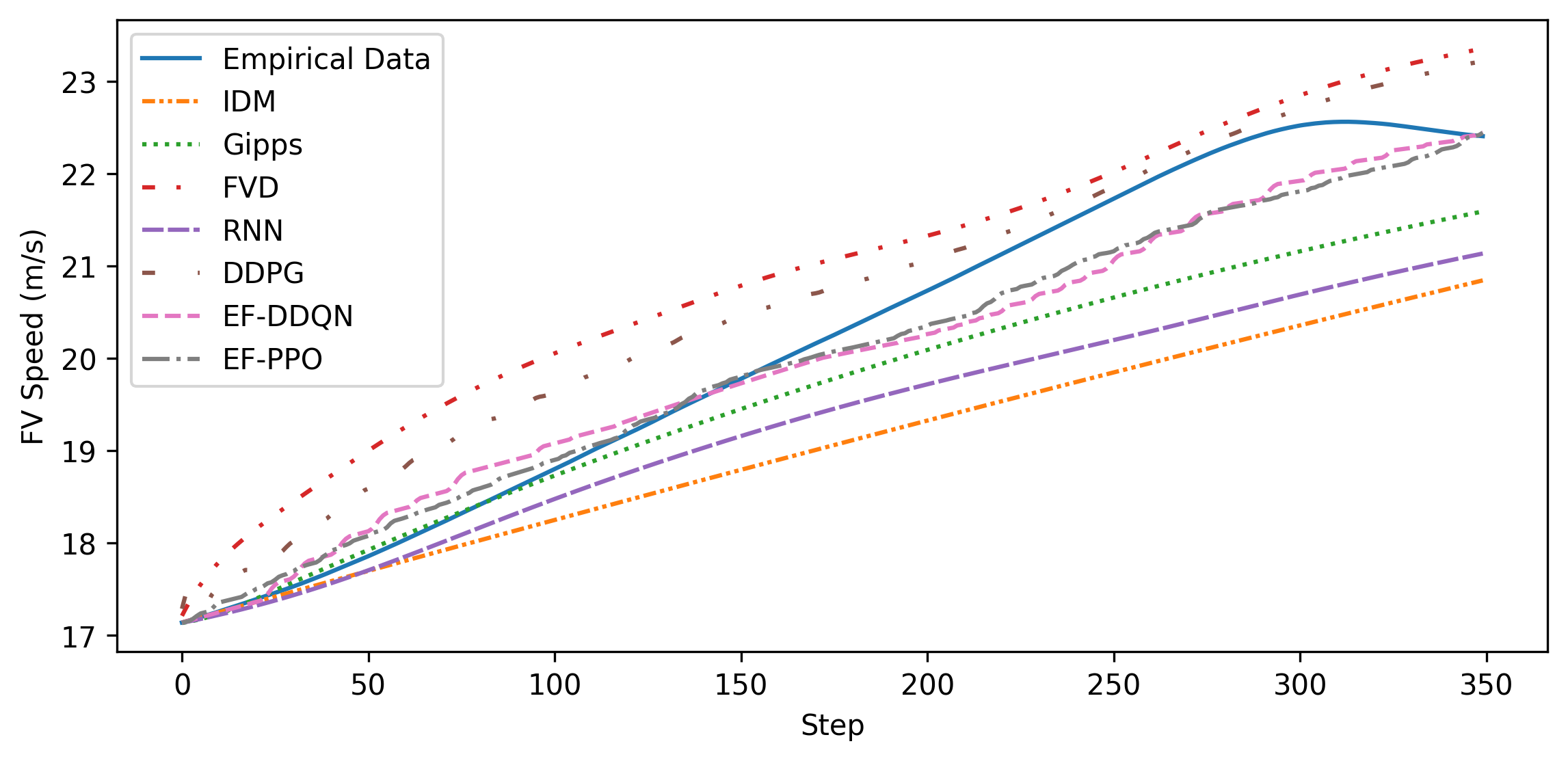}
\caption{Spacing and FV speed comparison for a case event.}
\label{Figure 5}
\end{figure}

\subsection{Capability of Integrating Hybrid Models}
To illustrate the details of what happened in the proposed EnsembleFollower framework, we visualized one representative car-following event outside the training set. Fig. \ref{Figure 5} shows the observed and simulated spacing and FV speed. It can be seen that the proposed models produced closer spacing and speed to the empirical data than the ingredient models did.

\begin{figure}
\centering
\includegraphics[width=3.4in]{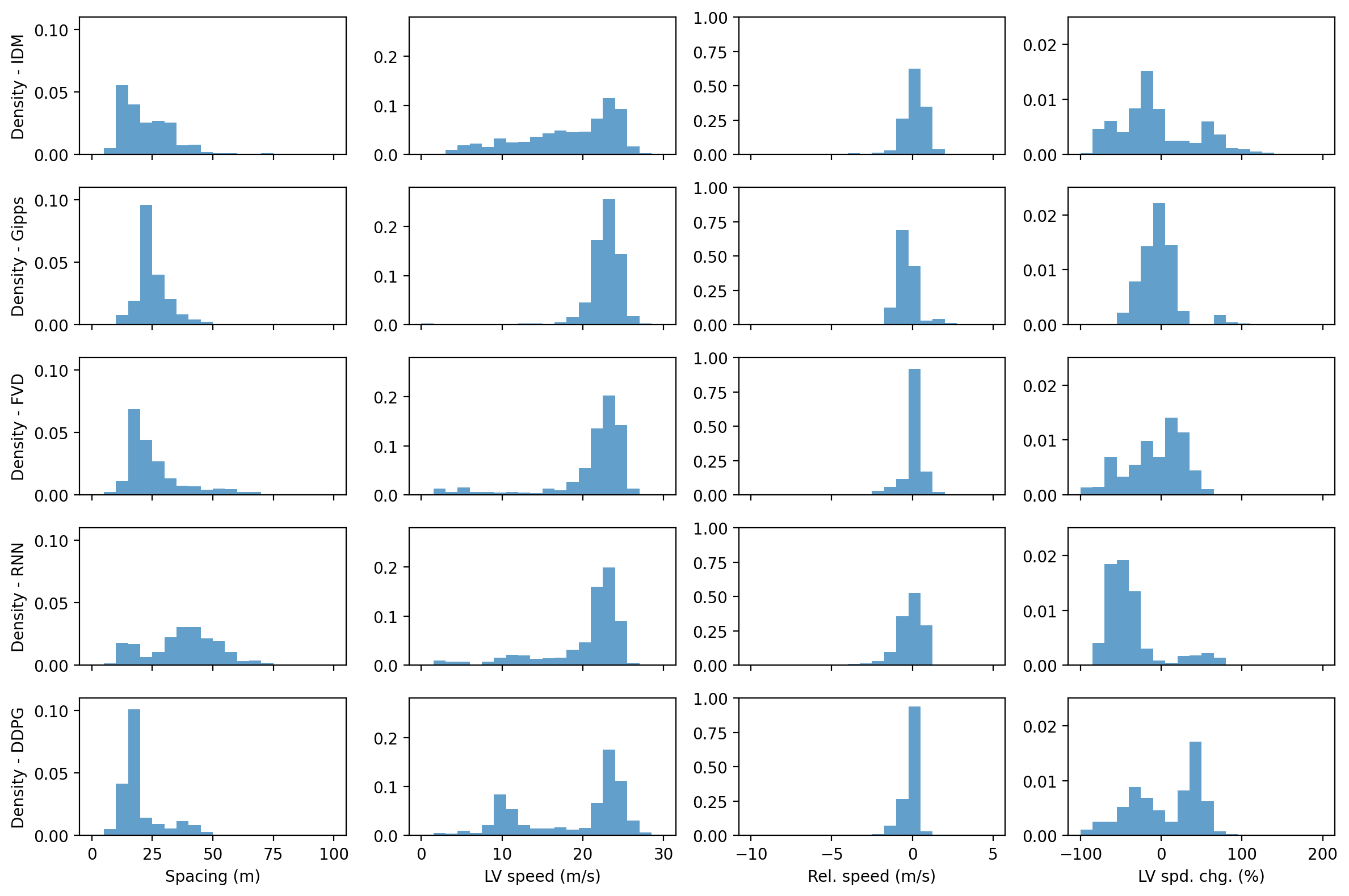}
\caption{EF-DDQN agent’s selections against driving states on the testing dataset.}
\label{dqn analysis}
\end{figure}

Furthermore, the results summarized in Table \ref{table:summary} and visualized in Fig. \ref{dqn analysis} indicate the frequency and context (aggregated in means and standard deviations) in which each low-level model was appointed by the EF-DDQN agent. The RNN and Gipps models were the most frequently selected, with over 30\% selection ratios respectively, while the other models also contributed notably. The preference for the RNN model occurred predominantly when the spacing between the LV and FV was large and the LV was braking over 1 second, potentially due to the RNN's sensitive reactions. In contrast, the DDPG model was often chosen when the LV had a relatively low speed and was starting to accelerate. The Gipps model was favored in moderate driving conditions, aligning with its foundation of maintaining safe distances. The IDM and FVD model appear to act as transitions among these varied operating states.

\begin{table}
\centering
\caption{Summary of the EF-PPO agent’s weight allocation on the testing dataset.}
\begin{tabular}{lllllll}
\hline
Model & Weight & Primary(\%) & Dominating(\%) \\ \hline
RNN & 0.477 ($\pm$ 0.158) & 56.29 & 34.83 \\
IDM & 0.353 ($\pm$ 0.156) & 39.52 & 10.20 \\
Gipps & 0.121 ($\pm$ 0.132) & 3.69 & 0.56 \\ 
DDPG & 0.035 ($\pm$ 0.075) & 0.35 & 0.07 \\
FVD & 0.015 ($\pm$ 0.049) & 0.15 & 0.02 \\ \hline
\end{tabular}
\label{table:summary2}
\end{table}

\begin{figure}
\centering
\includegraphics[width=3.4in]{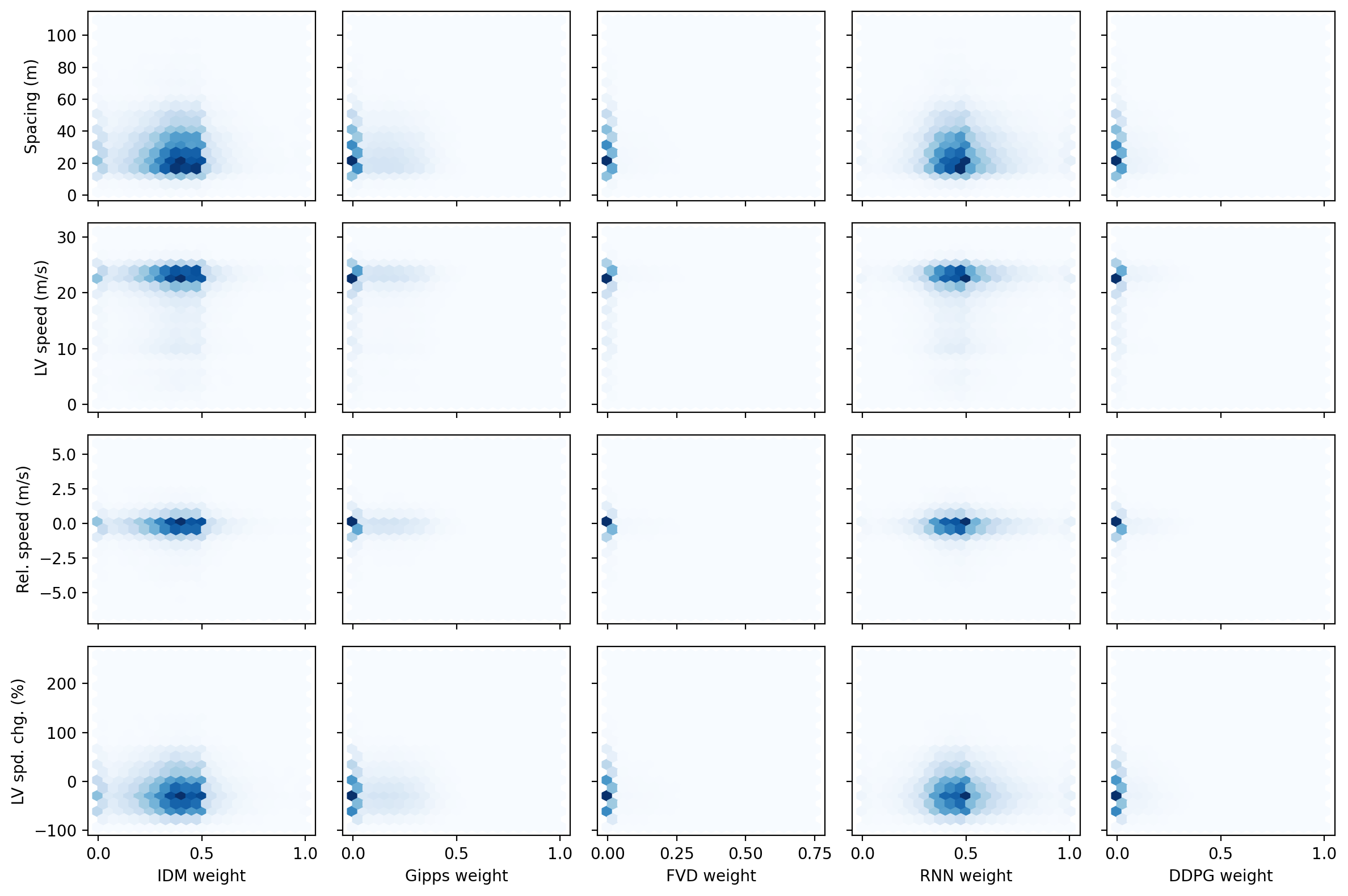}
\caption{EF-PPO agent’s weight allocation against driving states on the testing dataset.}
\label{ppo analysis}
\end{figure}

Fig. \ref{ppo analysis} illustrates the correlations between the weights assigned to each model by the EF-PPO agent and the corresponding driving states, with additional statistics in Table \ref{table:summary2}. The RNN method was the primary component of the model ensemble, obtaining the highest average weight and serving as the primary model in 56\% of cases. In approximately 35\% of cases, the RNN weight exceeded 0.5, indicating dominance. However, the risk of collisions with the original RNN model was mitigated by integrating other components, especially the IDM as the second main component known for robust performance. The minority models dominated in special cases as well, contributing to the generalization capability of the convex combination framework.

Overall, these findings demonstrate the capacity of the proposed approach to effectively manage diverse driving scenarios. By integrating hybrid complementary low-level models, the framework is able to leverage their respective strengths and compensate for individual weaknesses.

\section{Conclusion}\label{section:conclusion}
Motivated by the principles of EL, we present a novel method that integrates an ensemble of car-following models and RL techniques for autonomous vehicles. The contributions of this paper include the proposal of a hierarchical car-following framework called EnsembleFollower, and a comparison of EnsembleFollower with rule-based methods and data-driven models on real-world driving data. Through experiments on the HighD dataset, we have demonstrated that EnsembleFollower can reproduce accurate human-like driving behavior and outperform cutting-edge methods. EnsembleFollower effectively leverages the complementary strengths of hybrid car-following models in various situations, leading to accurate and efficient behavior planning. The proposed framework serves as an initial exemplar highlighting the benefits of integrating RL and EL techniques. Further research will be needed to fully explore the possibilities of this hybrid approach for enabling safe and efficient driving behavior.

A potential improvement to EnsembleFollower is adding more car-following models with diverse behavior styles as low-level components. Additionally, combining the spacing-based reward and the speed-based reward may contribute to better performance, which would need extensive experiments.

{\appendices
\section*{Appendix A}

Table \ref{low-calib} and Table \ref{low-train} are listed here for readers who are interested in the deployment details of the low-level car-following models.

\begin{table*}
\centering
\caption{\label{low-calib}Summary of the rule-based car-following model parameters and estimates.}
\begin{tabular}{lllll}
\hline
Model & Parameter (unit) & Short description & Bounds & Estimate \\ \hline
Gipps & $\widetilde{a}_n\left(\mathrm{~m} / \mathrm{s}^2\right)$ & Maximum desired acceleration of FV & {$[0.1,5]$} & 0.73 \\
 & $\tilde{b}_n\left(\mathrm{~m} / \mathrm{s}^2\right)$ & Maximum desired deceleration of FV & {$[0.1,5]$} & 2.30 \\
 & $\S_{n-1}(\mathrm{~m})$ & Effective length of LV & {$[5,15]$} & 6.96 \\
 & $\hat{b}\left(\mathrm{~m} / \mathrm{s}^2\right)$ & Maximum desired deceleration of LV & {$[0.1,5]$} & 1.92 \\
 & $\widetilde{V}_n(\mathrm{~km} / \mathrm{h})$ & Desired speed of FV & {$[1,150]$} & 24.52 \\
 & $\tau_n(\mathrm{~s})$ & Reaction time & {$[0.3,3]$} & 1.00 \\ \hline
IDM & $a_{\max }^{(n)}\left(\mathrm{m} / \mathrm{s}^2\right)$ & Maximum acceleration/deceleration of FV & {$[0.1,5]$} & 0.36 \\
 & $\widetilde{V}_n(\mathrm{~km} / \mathrm{h})$ & Desired speed of FV & {$[1,150]$} & 32.91 \\
 & $\beta$ & Acceleration exponent & {$[1,10]$} & 2.47 \\
 & $a_{\text {comf }}^{(n)}\left(\mathrm{m} / \mathrm{s}^2\right)$ & Comfortable deceleration of FV & {$[0.1,5]$} & 0.55 \\
 & $S_{j a m}^{(n)}(\mathrm{m})$ & Gap at standstill & {$[0.1,10]$} & 2.55 \\
 & $\widetilde{T}_n(\mathrm{~s})$ & Desired time headway of FV & {$[0.1,5]$} & 0.60 \\ \hline
FVD & $\alpha$ & Constant sensitivity coefficient & {$[0.05,20]$} & 0.22 \\
 & $\lambda_0{ }$ & Sensitivity to relative speed & {$[0,3]$} & 2.37 \\
 & $V_o(\mathrm{~km} / \mathrm{h})$ & Desired speed of FV & {$[1,252]$} & 24.00 \\
 & $b$ & Interaction length & {$[0.1,100]$} & 2.95 \\
 & $\beta$ & Form factor & {$[0.1,10]$} & 4.48 \\
 & $S_c(\mathrm{~m})$ & Max following distance & [10 120] & 56.35 \\ \hline
\end{tabular}
\end{table*}

\begin{table*}
\centering
\caption{\label{low-train}Hyperparameters of the data-driven car-following models.}
\begin{tabular}{lllll}
\hline
Model & Hyperparameter & Short description & Value \\ \hline
RNN & LSTM layer size & Number of Long Short-Term Memory layers & 1 \\
 & Bidirectional & Whether bidirectional layer is used & False \\
 & Epoch & Number of training Epochs & 20 \\
 & Batch size & Number of training cases used by gradient descent & 128 \\
 & Learning rate & Learning rate used by the optimizer & 0.001 \\ \hline
DDPG & Gamma & discount factor & 0.96 \\
 & Learning rate & Learning rate used by the optimizer & 0.001 \\
 & Batch size & Number of training cases used by gradient descent & 256 \\
 & Training start & Number of environmental steps collected before training & 100,000 \\
 & Buffer size & Maximum number of training cases in replay buffer & 1,000,000 \\
 & Tau & Soft update ratio of the target networks & 0.005 \\
 & Training step & Total training steps & 1,000,000 \\ \hline
\end{tabular}
\end{table*}

}


\bibliographystyle{IEEEtran}
\bibliography{my}

\begin{IEEEbiography}[{\includegraphics[width=1in,height=1.25in,clip,keepaspectratio]{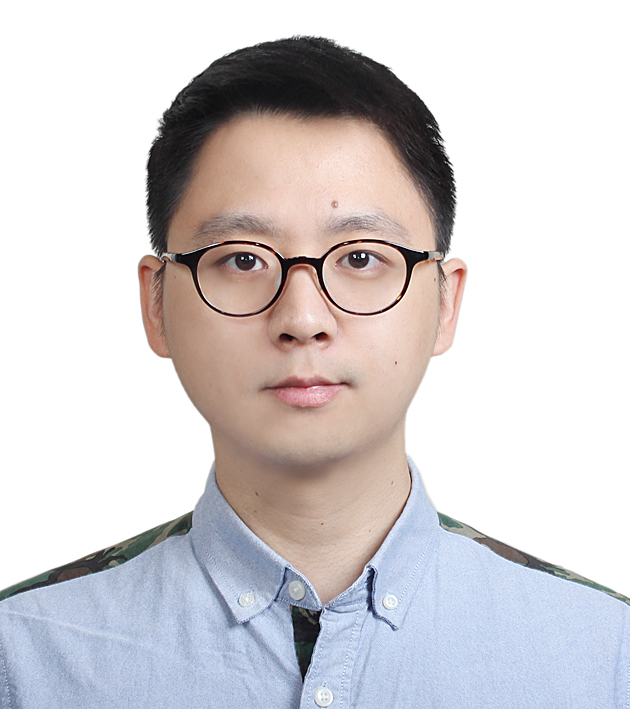}}]{Xu Han} received the M.Phil. degree in Internet of Things from the Hong Kong University of Science and Technology, Hong Kong, China, in 2015. He is currently pursuing the Ph.D. degree in Data Science and Analytics with the Hong Kong University of Science and Technology (Guangzhou), Guangzhou, China. His research interests include deep learning, reinforcement learning, decision intelligence and autonomous vehicles.
\end{IEEEbiography}

\begin{IEEEbiography}[{\includegraphics[width=1in,height=1.25in,clip,keepaspectratio]{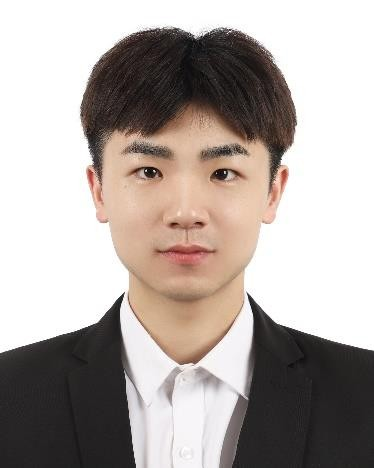}}]{Xianda Chen} received the M.S. degree from The University of Hong Kong, Hong Kong, China, in 2022. He is currently pursuing
the Ph.D. degree in intelligent transportation with The Hong Kong University of Science and Technology (Guangzhou), Guangzhou, China. His research interests include intelligent transportation, machine learning, and data analytics.
\end{IEEEbiography}

\begin{IEEEbiography}[{\includegraphics[width=1in,height=1.25in,clip,keepaspectratio]{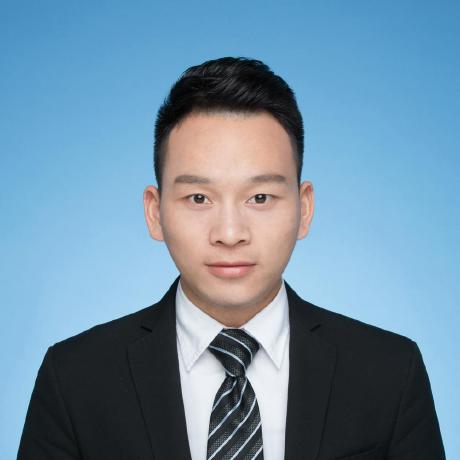}}]{Meixin Zhu} is a tenure-track Assistant Professor
in the Thrust of Intelligent Transportation (INTR)
under the Systems Hub at the Hong Kong University
of Science and Technology (Guangzhou). He is also
an affiliated Assistant Professor in the Civil and
Environmental Engineering Department at the Hong
Kong University of Science and Technology. He
obtained a Ph.D. degree in intelligent transportation
at the University of Washington (UW) in 2022. He
received his BS and MS degrees in traffic engineering in 2015 and 2018, respectively, from Tongji
University. His research interests include Autonomous Driving Decision
Making and Planning, Driving Behavior Modeling, Traffic-Flow Modeling
and Simulation, Traffic Signal Control, and (Multi-Agent) Reinforcement
Learning. He is a recipient of the TRB Best Dissertation Award (AED50)
in 2023.
\end{IEEEbiography}

\begin{IEEEbiography}[{\includegraphics[width=1in,height=1.25in,clip,keepaspectratio]{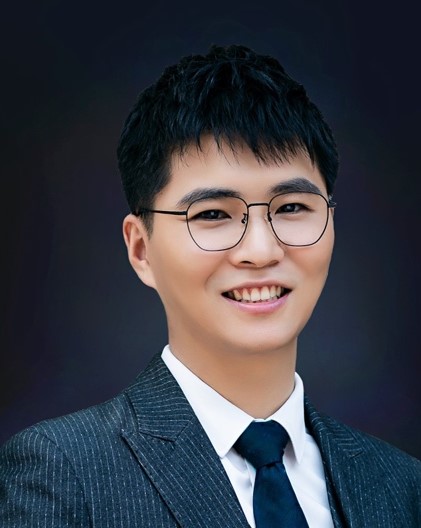}}]{Pinlong Cai} received his Ph.D. degree in Traffic Information Engineering and Control from Beihang University, Beijing, China. From 2016 to 2017, he was a Research Assistant with the Quanzhou Institute of Equipment Manufacturing, Haixi Institutes, Chinese Academy of Sciences. From 2020 to 2021, he was a Standard \& Strategy Engineer at ZTE Corporation. He is currently a Research Scientist at Shanghai Artificial Intelligence Laboratory. He was the winner of the Yangfan Special Project of the Shanghai Qimingxing Program in 2022. His research interests include data modelling, decision intelligence, autonomous vehicles, and cooperative vehicle infrastructure system.
\end{IEEEbiography}

\begin{IEEEbiography}[{\includegraphics[width=1in,height=1.25in,clip,keepaspectratio]{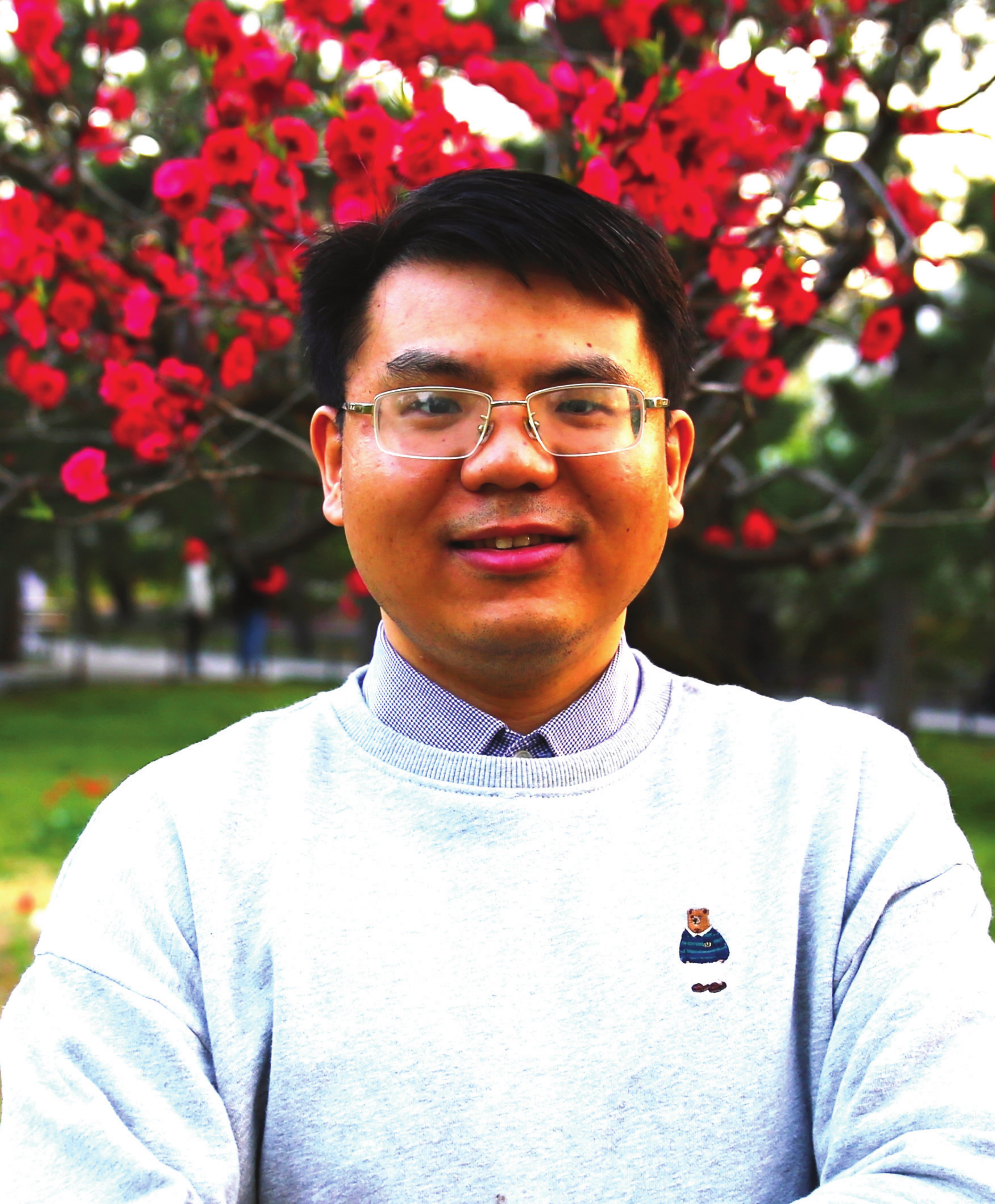}}]{Jianshan Zhou} received the B.Sc., M.Sc., and Ph.D. degrees in traffic information engineering and control from Beihang University, Beijing, China, in 2013, 2016, and 2020, respectively. He is an associate professor with the school of transportation science and engineering at Beihang University. From 2017 to 2018, he was a Visiting Research Fellow with the School of Informatics and Engineering, University of Sussex, Brighton, U.K. He was a Postdoctoral Research Fellow supported by the Zhuoyue Program of Beihang University and the National Postdoctoral Program for Innovative Talents from 2020 to 2022. He is or was the Technical Program Session Chair with the IEEE EDGE 2020, the IEEE ICUS 2022, the ICAUS 2022, the TPC member with the IEEE VTC2021-Fall track, and the Youth Editorial Board Member of the Unmanned Systems Technology. He is the author or co-author of more than 30 international scientific publications. His research interests include the modeling and optimization of vehicular communication networks and air–ground cooperative networks, the analysis and control of connected autonomous vehicles, and intelligent transportation systems. He was the recipient of the First Prize in the Science and Technology Award from the China Intelligent Transportation Systems Association in 2017, the First Prize in the Innovation and Development Award from the China Association of Productivity Promotion Centers in 2020, the Second Prize in the Beijing Science and Technology Progress Award in 2022, the National Scholarships in 2017 and 2019, the Outstanding Top-Ten Ph.D. Candidate Prize from Beihang University in 2018, the Outstanding China-SAE Doctoral Dissertation Award in 2020, and the Excellent Doctoral Dissertation Award from Beihang University in 2021.
\end{IEEEbiography}

\begin{IEEEbiography}[{\includegraphics[width=1in,height=1.25in,clip,keepaspectratio]{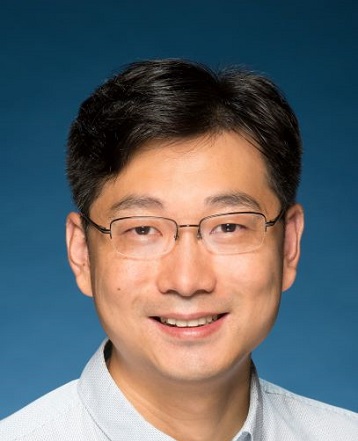}}]{Xiaowen Chu}
(xwchu@ust.hk) is currently a Professor at the Data Science and Analytics Thrust, The Hong Kong University of Science and Technology (Guangzhou). Dr. Chu received his B.Eng. degree in Computer Science from Tsinghua University in 1999, and the Ph.D. degree in Computer Science from HKUST in 2003. He has been serving as the associate editor or guest editor of many international journals, including IEEE Transactions on Network Science and Engineering, IEEE Internet of Things Journal, IEEE Network, and IEEE Transactions on Industrial Informatics. He is a co-recipient of the Best Paper Award of IEEE INFOCOM 2021. His current research interests include GPU Computing, Distributed Machine Learning, and Wireless Networks.
\end{IEEEbiography}



\vfill

\end{document}